\pdfoutput=1

\documentclass[11pt]{article}

\usepackage[final]{acl}

\usepackage{times}

\usepackage{microtype}
\usepackage{hyperref}
\usepackage{url}

\usepackage{lineno}

\definecolor{darkblue}{rgb}{0, 0, 0.5}
\hypersetup{colorlinks=true, citecolor=darkblue, linkcolor=darkblue, urlcolor=darkblue}

\usepackage{inconsolata}

\usepackage{graphicx}

\usepackage[]{algorithm2e}
\usepackage{algorithmic,float}
\usepackage[utf8]{inputenc} 
\usepackage[T1]{fontenc}    
\usepackage{booktabs}       
\usepackage{rotating}       
\usepackage{amsfonts}       
\usepackage{nicefrac}       
\usepackage{microtype}      
\usepackage{multirow}
\usepackage{caption}
\usepackage{amsmath,amssymb}
\usepackage{dsfont}			
\usepackage{cases}
\usepackage{color}
\usepackage{tikz}
\usepackage{pgfplots, pgfplotstable}
\usetikzlibrary{shapes,arrows,calc}
\usepackage{caption}
\usepackage{textcomp}
\usepackage{subcaption}
\usepackage{enumitem}
\usepackage{import}
\usepackage{CJKutf8}
\usepackage[toc,page]{appendix}
\usepackage[textsize=large]{todonotes}
\usepackage[normalem]{ulem}
\usepackage[scaled=.8]{beramono}
\usepackage{fancyvrb}
\newcommand{\hochkomma}{$^{,}$}
\useunder{\uline}{\ul}{}

\setlist[itemize]{leftmargin=15pt}

\newcommand\blfootnote[1]{%
  \begingroup
  \renewcommand\thefootnote{}\footnote{#1}%
  \addtocounter{footnote}{-1}%
  \endgroup
}
\newcommand{\textcite}[1]{\citeauthor{#1}, \citeyear{#1}}
\title{Extending Automatic Machine Translation Evaluation to\\ Book-Length Documents}

\author{Kuang-Da Wang$^1\dagger$, Shuoyang Ding$^2\dagger$, Chao-Han Huck Yang$^2$,\\ \textbf{Ping-Chun Hsieh$^1$, Wen-Chih Peng$^1$, Vitaly Lavrukhin$^2$, Boris Ginsburg$^2$} \\
$^1$National Yang Ming Chiao Tung University, $^2$NVIDIA \\
\texttt{gdwang.cs10@nycu.edu.tw\quad shuoyangd@nvidia.com} \\
}

\begin{document}

\maketitle

\begin{abstract}
Despite Large Language Models (LLMs) demonstrating superior translation performance and long-context capabilities, evaluation methodologies remain constrained to sentence-level assessment due to dataset limitations, token number restrictions in metrics, and rigid sentence boundary requirements.
We introduce SEGALE, an evaluation scheme that extends existing automatic metrics to long-document translation by treating documents as continuous text and applying sentence segmentation and alignment methods.
Our approach enables previously unattainable document-level evaluation, handling translations of arbitrary length generated with document-level prompts while accounting for under-/over-translations and varied sentence boundaries.
Experiments show our scheme significantly outperforms existing long-form document evaluation schemes, while being comparable to evaluations performed with groundtruth sentence alignments. 
Additionally, we apply our scheme to book-length texts and newly demonstrate that many open-weight LLMs fail to effectively translate documents at their reported maximum context lengths.
\end{abstract}
\blfootnote{$\dagger$ Equal contribution.}

\section{Introduction}
Since the inception of Large Language Models (LLMs), the paradigm of machine translation (MT) has been shifting toward an LLM-based approach.
In the WMT 2024 general translation shared task \citep{kocmi-etal-2024-findings}, LLM-based systems demonstrated strong performance, ultimately dominating submissions across all language pairs.
Additionally, because of their long context windows, LLM-based systems may potentially be able to generate translations that better capture discourse-level phenomena and maintain coherence across longer spans of text.
This development aligns with the long-standing trend in MT research to move beyond sentence-level processing toward \emph{paragraph-level} \citep{deutsch-etal-2023-training}, \emph{discourse-level} \citep{bawden-etal-2018-evaluating}, and \emph{document-level} \citep{zhu-etal-2024-preference} translation.

However, despite claims that modern LLMs can process inputs of up to 1M tokens \citep{DBLP:journals/corr/abs-2501-15383}, evaluations of LLM-based translations remain largely confined to \emph{sentence-level} or \emph{segment-level}, in that they are only prompted to translate one sentence or segment at a time.
This forms a significant gap between what LLMs can generate and what existing metrics can evaluate.
This limitation stems from the following two challenges:
\begin{enumerate}[leftmargin=*] \itemsep -3pt
    \item Commonly used model-based evaluation metrics have relatively low maximum token length limitations (e.g., 512 tokens for COMET).
    \item Current automatic evaluation metrics require adherence to pre-defined sentence boundaries. This forces evaluators to either use sentence-level prompting or add artificial boundaries.
\end{enumerate}


In this paper, we propose SEGALE (\uline{SEG}ment, \uline{AL}ign, and \uline{E}valuate), a scheme that solves these challenges by extending existing automatic evaluation metrics to long documents.
To summarize, our approach applies to arbitrarily long documents by using sentence segmenters and aligners to create appropriate sentence-level alignments.
We treat those automatically aligned sentence pairs in two different ways.
In the case where a valid sentence alignment is found, we apply the existing evaluation metric to the sentence pair.
In the case where translation errors occur - either when content from the source text is missing in the translation (\emph{under-translation}) or when there is hallucinated content that is not present in the source (\emph{over-translation}) - we detect these as null alignments and assign a fixed penalty.
At the end, along with metric scores, we also report the ratio of null alignments (NA ratio) to help track these over-translation and under-translation errors, which current MT evaluation metrics are having trouble detecting reliably.

Our experiments demonstrate that this scheme evaluates translations with comparable performance to existing sentence-level metrics when applied to cases with over-translation and under-translation.
In addition, it handles cases where LLMs are liberal with sentence boundaries, which create many-to-one and one-to-many sentence alignments.
Lastly, we newly demonstrate that we can successfully apply this scheme to evaluate translations of book-length texts, and reveal that many open-weight LLMs cannot translate documents of their reported context length, because the number of under- and over-translation errors rise sharply as the input length gets longer.
Our code and artifacts will be available at \url{https://github.com/nvlabs/SEGALE}.

\section{Related Work}

\subsection{Document-Level Translation}

Document-level MT extends translation beyond isolated sentences by leveraging broader context for coherence.
Existing approaches include simply concatenating adjacent sentences as a larger input to a standard MT model \citep{scherrer-etal-2019-analysing,junczys-dowmunt-2019-microsoft,sun-etal-2022-rethinking},
as well as more advanced architectures that introduce context-specific modules: multi-encoder models encode previous sentences with separate encoders and hybrid attention mechanisms \citep{DBLP:journals/corr/JeanLFC17,bawden-etal-2018-evaluating,voita-etal-2019-good,miculicich-etal-2018-document,maruf-etal-2019-selective,herold-ney-2023-improving}.
Recent work has also focused on improving the quality of document-level translation by utilizing larger-scale document-level corpus \citep{thai-etal-2022-exploring, al-ghussin-etal-2023-exploring, DBLP:journals/corr/abs-2304-12959,pal-etal-2024-document}, as well as leveraging large language models (LLMs) \citep{karpinska-iyyer-2023-large, wang-etal-2023-document-level}.

Despite the progress, document-level translation still has a few limitations. 
First, a lot of work stick to a relatively small number of maximum input/output length.
For example, \citet{scherrer-etal-2019-analysing,DBLP:journals/corr/abs-2304-12959} both have maximum context length of 250 tokens, while \citet{al-ghussin-etal-2023-exploring,pal-etal-2024-document} have 512.
Besides, some work \citep{junczys-dowmunt-2019-microsoft,DBLP:journals/corr/abs-2304-12959} introduce artificial sentence boundaries to the input, which provides native sentence segmentations for evaluation.
This requires specialized training data or prompt, and there is no guarantee that the system will generate matching sentence boundaries as the input document.

\subsection{Machine Translation Evaluation}

Machine translation evaluation has shifted from string-based metrics (e.g., BLEU \citep{papineni-etal-2002-bleu}, chrF \citep{popovic-2015-chrf}) to model-based metrics (e.g., COMET \citep{rei-etal-2020-comet}, MetricX \citep{juraska-etal-2024-metricx}, GEMBA \citep{kocmi-federmann-2023-large}).
Human evaluations like direct assessment (DA) and multi-dimensional quality metrics (MQM) played a crucial role in this paradigm shift by providing meta-evaluations and training data for model-based metrics.

Most model-based metrics are trained and evaluated on the segment-level.
For example, COMET limits each input (source, target, and reference) to 512 tokens, while MetricX has a combined limit of 1,536 tokens across all inputs.
In contrast, Qwen-2.5 \cite{qwen2-5}, a recent open-source LLM, can handle input of 131,072 tokens and generate up to 8,192 tokens.
Prior efforts have explored extending MT evaluation metrics beyond sentence-level.
\citet{vernikos-etal-2022-embarrassingly} proposed adding prior sentences as context when training model-based metrics.
\citet{deutsch-etal-2023-training} trained metrics on paragraph-level data but found limited benefits.
These studies are orthogonal to ours -- they focus on building new model-based metrics with longer maximum input length, while we focus on applying existing metrics to long-form text.

Closest to the spirit of our work is mwerSegmenter ~\citep{DBLP:conf/iwslt/MatusovLBN05}. It is a joint sentence segmentation and alignment scheme that has been the long-standing evaluation standard for unsegmented speech translation\footnote{Specifically, mwerSegmenter has been the evaluation standard for the IWSLT speech translation shared tasks (\url{https://iwslt.org/})}.
The high-level idea is to jointly segment and align long-form model output by minimizing the word error rate (WER) between the segmented text and the already-segmented reference text.
The assumption behind the idea is that perfectly segmented and aligned sentences are more likely to be translated well, and thus should have a low WER.
Similar to mwerSegmenter, \citet{wang-etal-2023-document-level} implemented a segmentation and alignment scheme based on Bleualign \citep{DBLP:conf/amta/SennrichV10}, but there was no extensive discussion regarding the validity of the scheme.
Apart from that, \citet{raunak-etal-2024-slide} proposed to extend existing metrics based on running evaluations on aligned sliding windows over sentences in a document, but the algorithm is still limited to the sentence-level prompting paradigm.

A few recent investigations \citep{salesky-etal-2023-evaluating,sperber-etal-2024-evaluating} of mwerSegmenter in the context of long-form audio data raised concerns about the segmentation quality.
The reader shall see that our results corroborate the concerns.
Contemporary to our work, \citet{post-hoang-2025-effects} resolved a few issues with the current mwerSegmenter implementation such as lack of standardized tokenization and empty translation hypotheses.


\subsection{Long-Context LLM Evaluation}

Recent progress in extending LLM architectures to handle longer contexts has spurred considerable research interest.
Parallel to architectural advances, there has been growing attention toward systematically evaluating the capabilities of LLMs on long-context tasks.
\citet{kamradt2023niah} developed an evaluation focusing on models' abilities to retrieve deeply nested information.
Similarly, \citet{bai-etal-2024-longbench} introduced a long-context bilingual benchmark for assessing models' comprehension and reasoning abilities, 
while \citet{an-etal-2024-l} shows that standardized evaluation criteria across multiple long-context scenarios are essential for comprehensive model assessment.
Furthermore, \citet{DBLP:journals/corr/abs-2404-06654} highlights that there are discrepancies between theoretical capabilities and effective usable context lengths of contemporary LLMs.

Despite these advances, the evaluation methodologies have predominantly focused on general comprehension tasks rather than specialized applications like long-context machine translation. Existing metrics face limitations such as fixed maximum token lengths and rigid assumptions about sentence boundaries, which hinder effective evaluation of extensive, continuous texts, like books.

\section{Preliminaries} \label{sec:preliminaries}

\begin{figure}
  \centering
  \includegraphics[width=\columnwidth]{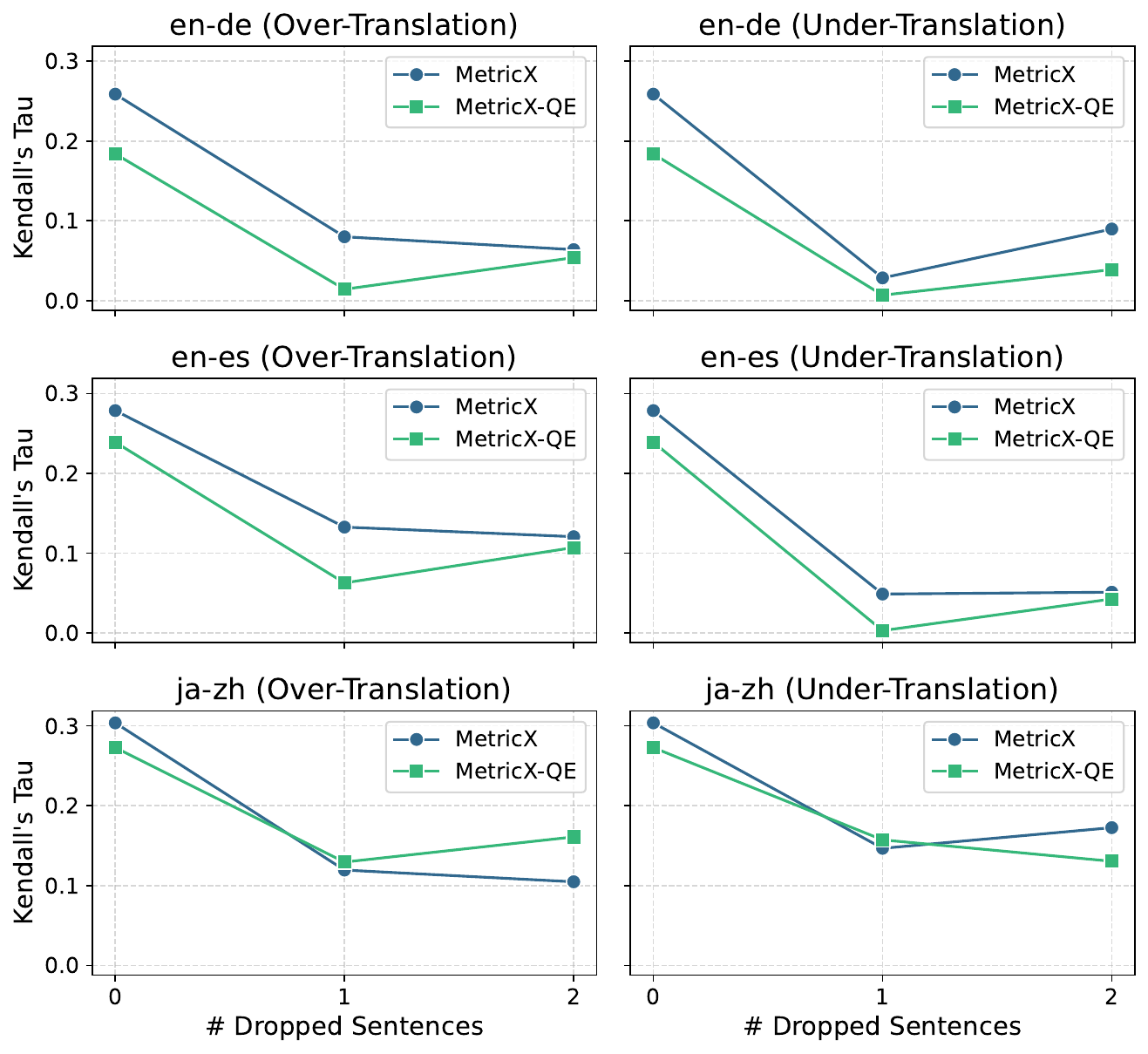}
  \caption{The Kendall’s $\tau$ correlations of MetricX-24 and MetricX-24-QE show limited sensitivity to over- and under-translation; sentences with more than three drops are insufficient to estimate correlations reliably.}
  \label{fig:metricx_metaeval}
\end{figure}

\begin{figure*}
  \centering
  \includegraphics[width=\textwidth]{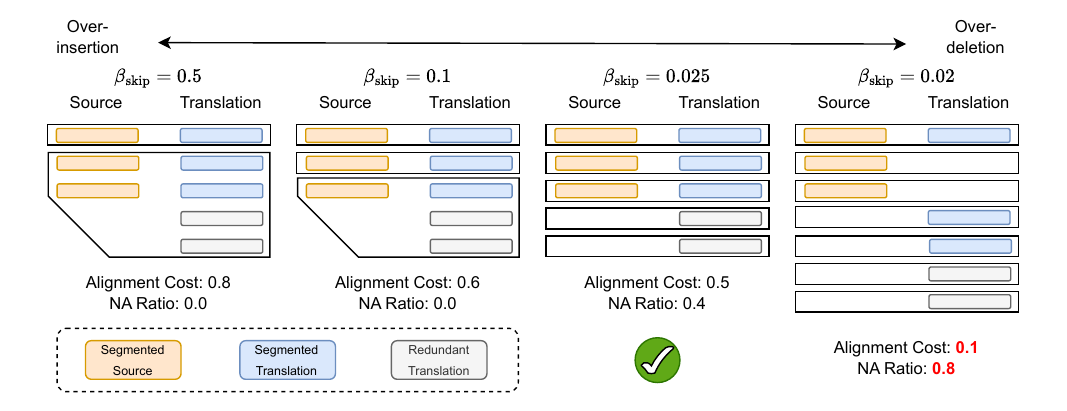}
  \caption{The effect of the skip cost (\( \beta_{\text{skip}} \)) on alignment behavior for over-translation. Higher skip costs increase the risk of over-insertion by allowing loose semantic matches to align, while lower skip costs enforce stricter alignment, leading to over-deletion. Over-deletion is indicated by spikes in the null alignment ratio (NA ratio) and low alignment costs, both shown in red.}
  \label{fig:sentence_alignment}
\end{figure*}
Our ultimate goal is to find a way to evaluate the translation quality in the following scenarios:

\begin{itemize} \itemsep -3pt
  \item For translations of documents of arbitrary length generated with document-level prompts, while handling under-/over-translations and varied sentence boundaries
  \item For both reference-based and reference-free evaluations, thus enabling broader applications like curating high-quality training data for LLMs (similar to \textcite{finkelstein-etal-2024-introducing})
\end{itemize}

%


We start by justifying why extensions of existing metrics are required for document-level MT evaluation, rather than directly feeding concatenated sentence pairs from a document into existing metrics.
A key limitation of the direct concatenation approach is that commonly used model-based evaluation metrics have relatively low maximum token length limits.
Additionally, even for documents that are within the maximum token length limit, we show with a preliminary study in this section that directly applying state-of-the-art MT metrics to concatenated sentences is actually not able to reliably detect under- and over-translation errors.\footnote{The conclusion may seem different from a prior study \citet{deutsch-etal-2023-training}, but it's actually not a direct contradiction, because the evaluation in \citet{deutsch-etal-2023-training} focuses only on cases where one-to-one mapping between source, target, and reference exists.}

We evaluate the performance of MetricX-24 \cite{juraska-etal-2024-metricx} on such concatenation approach.
To avoid going over the maximum token length limit of MetricX-24, we filter out cases where the concatenation of source and target inputs exceed 1024 tokens in length.
We compute both MetricX-24 and its reference-free variant, MetricX-24-QE, across three language pairs: English–German (en-de), English–Spanish (en-es), and Japanese–Chinese (ja-zh).
We use the dataset from the WMT 2024 Metrics Shared Task~\citep{freitag-etal-2024-llms}, which contains: (1) source texts and reference translations in WMT 2024 General Translation Shared Task, (2) translations of the source texts from the system submissions, and (3) MQM human evaluations of the system translations.

To simulate translation errors, we manipulate the texts in two ways: for over-translation, we remove one or two sentences from both source and reference texts, while for under-translation, we remove sentences only from the system translations.\footnote{We limit ourselves to removing at most two sentences since removing more would leave too few documents for reliable Kendall's $\tau$ correlation calculations.}
We measure the performance of the metrics by computing Kendall’s $\tau$ correlations between the metric outputs and human evaluation scores.\footnote{For more details on meta-evaluation, see Section~\ref{subsec:setup}.}

The results of this preliminary study (Figure~\ref{fig:metricx_metaeval}) confirm that MetricX-24 and MetricX-24-QE have limited sensitivity to over- and under-translations, even within token length limits.
This empirical evidence shows that the direct concatenation approach is inadequate for document-level MT evaluation.
To apply existing MT metrics to document-level MT evaluation, we need an extension scheme that can properly handle these translation errors while working within the constraints of existing metrics.


\section{Method}

We now introduce SEGALE, our proposed extension scheme. SEGALE consists of three steps:
\vspace{-0.2cm}
\begin{enumerate} \itemsep -3pt
    \item \uline{SEG}ment the document-level system translation into indivudual sentences
    \item \uline{AL}ign the source and system translations
    \item \uline{E}valuate the aligned sentence blocks with existing metrics, then average sentence-level scores to obtain a document-level score
\end{enumerate}

\subsection{Sentence Segmentation}

We use off-the-shelf sentence segmentation models to segment documents into sentences.
We experimented with ersatz \cite{wicks-post-2021-unified} and spaCy \cite{spacy}.

%

\subsection{Sentence Alignment}
Given a sentence-segmented source document \( \mathbf{S} = \{s_1, \ldots, s_N\} \) and its translation \( \mathbf{T} = \{t_1, \ldots, t_M\} \), the goal of sentence alignment is to identify a minimal-cost alignment path that maps contiguous spans of source sentences to contiguous spans of target sentences.
We use Vecalign~\citep{thompson-koehn-2019-vecalign} to perform such alignment, with a few changes detailed below.




\paragraph{Adaptive Penalty Search}

Ideally, all over- and under-translation errors will result in \emph{null alignments}.
In Vecalign, null alignments are modeled via a skip cost, which is parameterized by a percentile-based threshold \( \beta_{\text{skip}} \).
If the skip cost is set too high, many alignments are forced between unrelated sentence blocks -- essentially reverting to the scenario in our preliminary experiment.
Conversely, if the skip cost becomes too low, the aligner will start to assign null alignments even to semantically related sentence pairs.
Figure~\ref{fig:sentence_alignment} illustrates an example of over-translation and demonstrates how different \(\beta_{\text{skip}}\) values impact the alignment result.

Given that the optimal value of \(\beta_{\text{skip}}\) can vary depending on the severity of over- or under-translation in each individual document, we implement an adaptive search strategy to enhance robustness.
We leverage the insight that over-deletion is often signaled by sudden spikes in the null alignment ratio and abnormally low average alignment costs. 
Since the optimal alignment typically occurs just before over-deletion sets in, our approach starts with a relatively high \(\beta_{\text{skip}}\) and progressively decreases it in small steps. 

At each step, alignment quality is monitored using two heuristics to determine whether to terminate the search: (a) when the average alignment cost drops below a threshold, indicating excessive skipping, or (b) when the null alignment ratio exceeds a predefined limit at a step. Both patterns typically suggest that the skip penalty has become too lenient.
In such cases, we revert to the previous step and treat it as the final alignment result.
See Appendix~\ref{app:imple_details} for implementation details and heuristic settings.

\paragraph{Building Better Text Embeddings}

Text embedding models are crucial for sentence alignment.
We observe that sentence segmentation granularities vary across languages, which strains existing text embedding models.
For example, suppose we have a long source sentence $s$ that should align to smaller target sentences $\{t_1, \ldots, t_M\}$.
In Vecalign, the scoring function calculates similarity between $s$ and all consecutive blocks of $\{t_1, \ldots, t_M\}$.
This is not what text embedding models are trained for, leading to suboptimal alignments.

Motivated by this observation, we build our own text embedding model that is specifically designed to handle the sentence segmentation granularities we described above.
Our model is fine-tuned from BGE-M3 \cite{bge-m3}, which achieves high performance on bitext-mining task with only 568M parameters and without relying on instructions.
The fine-tuning is performed on a synthetic dataset with query, positive and negative example triplets built from the News Commentary v18\footnote{\url{https://data.statmt.org/news-commentary/v18.1/}} dataset (see more details in Appendix~\ref{subsec:emb_finetune}), with the FlagEmbedding toolkit\footnote{\url{https://github.com/FlagOpen/FlagEmbedding}}.
The readers shall see that our fine-tuned text embedding model outperforms both LASER~\citep{artetxe-schwenk-2019-massively} and BGE-M3 text embedding models in our experiments.

\subsection{Evaluation via Existing Metrics}

Once the target translation is segmented and aligned with the source document, we calculate the segment-level translation quality using existing metrics, then average the scores to obtain a document-level score.
In the case when a non-one-to-one alignment is established, we concatenate the aligned sentence blocks and evaluate them with the metric.
Two numbers are reported per document: the average segment score and ratio of null alignments over aligned sentence pairs ("NA ratio").
For each null alignment, we assign the worst possible score (0 for COMET, 25 for MetricX) and include it in the average calculation.
\begin{table*}[h]
    \centering
    \scalebox{0.85}{
        \begin{tabular}{@{}lrrrrr@{\hspace*{2pt}}l@{}}
            \toprule
                                          & \multicolumn{1}{c}{\textbf{COMET}} & \multicolumn{1}{c}{\textbf{COMET-QE}} & \multicolumn{1}{c}{\textbf{MetricX}} & \multicolumn{1}{c}{\textbf{MetricX-QE}} & \multicolumn{2}{c}{\textbf{NA Ratio ($\left|\Delta_\text{Gold}\right|$)}} \\ \midrule
            \textbf{Original}             & \multicolumn{1}{l}{}               & \multicolumn{1}{l}{}                  & \multicolumn{1}{l}{}                 & \multicolumn{1}{l}{}                    & \multicolumn{1}{l}{}          &                       \\
            Gold                          & 0.3110                             & 0.2800                                & 0.3085                               & 0.2683                                  & \quad\quad\,\, 0.0\%          & (--)                  \\
            SEGALE                        & 0.3085                             & 0.2768                                & 0.3074                               & 0.2630                                  & 0.7\%                         & (0.7\%)               \\
            mwerSegmenter                 & 0.2874                             & 0.2547                                & 0.2799                               & 0.2360                                  & 0.0\%                         & (0.0\%)               \\ \midrule
            \textbf{Over-Translate}       & \multicolumn{1}{l}{}               & \multicolumn{1}{l}{}                  & \multicolumn{1}{l}{}                 & \multicolumn{1}{l}{}                    & \multicolumn{1}{l}{}          &                       \\
            Gold                          & 0.3848                             & 0.3547                                & 0.3598                               & 0.3001                                  & 10.0\%                        & (--)                  \\
            SEGALE                        & 0.3532                             & 0.3368                                & 0.3499                               & 0.3082                                  & 11.2\%                        & (1.2\%)               \\
            mwerSegmenter                 & 0.3506                             & 0.3213                                & 0.3251                               & 0.2729                                  & 0.0\%                         & (10.0\%)              \\ \midrule
            \textbf{Under-Translate}      & \multicolumn{1}{l}{}               & \multicolumn{1}{l}{}                  & \multicolumn{1}{l}{}                 & \multicolumn{1}{l}{}                    & \multicolumn{1}{l}{}          &                       \\
            Gold                          & 0.3521                             & 0.3174                                & 0.3102                               & 0.2832                                  & 10.0\%                        & (--)                  \\
            SEGALE                        & 0.3493                             & 0.3295                                & 0.3268                               & 0.2852                                  & 5.8\%                         & (4.2\%)               \\
            mwerSegmenter                 & 0.2183                             & 0.1903                                & 0.1770                               & 0.1441                                  & 2.1\%                         & (7.9\%)               \\ \midrule
            \textbf{Flex-Boundary}        & \multicolumn{1}{l}{}               & \multicolumn{1}{l}{}                  & \multicolumn{1}{l}{}                 & \multicolumn{1}{l}{}                    & \multicolumn{1}{l}{}          &                       \\
            Gold                          & 0.3096                             & 0.2788                                & 0.3066                               & 0.2665                                  & 0.0\%                         & (--)                  \\
            SEGALE                        & 0.3067                             & 0.2746                                & 0.3033                               & 0.2597                                  & 1.2\%                         & (1.2\%)               \\
            mwerSegmenter                 & 0.2611                             & 0.2325                                & 0.2524                               & 0.2131                                  & 0.0\%                         & (0.0\%)               \\ \bottomrule
        \end{tabular} 
    }
    \caption{
        Correlation between document-level scores and human judgments under different evaluation settings.
        The first four columns are Kendall's $\tau$ correlation coefficients ($\uparrow$), with the last column being the average NA ratio and the absolute difference from the groundtruth ($\left|\Delta_\text{Gold}\right|$).
        All numbers are averaged across three language pairs (en-de, en-es, ja-zh),
        and all reported numbers of our method are calculated with ersatz sentence segmenter and our fine-tuned BGE-M3 text embedding model.
        }
    \label{tab:main}
\end{table*}

\section{Experiments}


\subsection{Setup} \label{subsec:setup}

Our experiments are aimed to test if the proposed evaluation scheme can achieve the two goals stated in Section~\ref{sec:preliminaries} -- in other words, whether it is (1) robust to all kinds of anomalies in system translations and (2) effective with both reference-based and reference-free metrics.
Our data and metric setup reflect the above goals.

To establish meaningful comparisons, we compare with two baselines.
One is calculating metric scores using the groundtruth sentence boundaries and alignments provided by the dataset ("Gold"), which serves as a performance upper bound.\footnote{As the reader shall see, there are times when our score is higher than the upper bound performance. This is likely caused by the sentence segmentation variations between the sentence segmenter and boundaries in the test set. It shouldn't be interpreted as our method being better in a meaningful way.}
The other is calculating metric scores using the sentence boundaries and alignments derived by mwerSegmenter \cite{DBLP:conf/iwslt/MatusovLBN05}.

\paragraph{Dataset}
We use the same dataset from preliminary experiments in Section~\ref{sec:preliminaries}.
In all experiments, we merge existing sentence boundaries in the system translation to simulate system translations generated at document-level.
We adhere to the same sentence boundaries on the source and reference sides during evaluation.

There are significant limitations if we only conduct meta-evaluation on the original test set, because the original test set is always guaranteed to have perfect sentence alignments (i.e., no null alignments).
Hence, in addition to the original test set (\textit{"original"} case), we create three synthetic test sets by introducing anomalies into the original test set, namely \textit{"over-translate"}, \textit{"under-translate"}, and \textit{"flex-boundary"} cases.
The first two cases are created by randomly removing 10\% of the sentences from the source/reference sides and system translations, respectively.
The last case is created by merging 10\% of neighboring sentences in the source texts with GPT-4o, using carefully designed prompts to preserve the original semantic content and word choices.
Since the system translations were generated from the source texts before merging, this introduces sentence boundary variations without modifying the system translations and their accompanying human judgments.
For more details, please refer to Appendix~\ref{app:merge_case}.




\paragraph{Underlying Metrics}
Our experiments cover both reference-based and reference-free ("QE") variants of COMET{\footnote{Reference-based: \href{https://huggingface.co/Unbabel/wmt22-comet-da}{\texttt{wmt22-comet-da}}. Reference-free: \href{https://huggingface.co/Unbabel/wmt22-cometkiwi-da}{\texttt{wmt22-cometwiki-da}}} and MetricX{\footnote{\href{https://huggingface.co/google/metricx-24-hybrid-large-v2p6}{\texttt{metricx-24-hybrid-large-v2p6}}}}.

\paragraph{Meta-Evaluation}


Similar to previous work and preliminary experiments, we use correlation between document-level scores and human judgments as the primary metric.
Although both system translation and human judgments are performed at segment-level, previous work~\cite{deutsch-etal-2023-training} has shown that MQM annotations are done with context of surrounding sentences, and sentences appear in document order.
Hence, they are a good proxy for document translation quality.
For cases with introduced null alignments, we assign $25$ as the human-annotated MQM score for each null alignment, which is then converted into z-score in accordance with each human annotator's scoring distribution.
Like previous work, we average the segment-level human judgment scores as the document-level scores.

We also report NA ratio for each method as the auxiliary metric. Ideally, we would like to achieve the same NA ratio as the groundtruth ($\left|\Delta_\text{Gold}\right| = 0$), but the reader should note that perfect NA ratio on its own doesn't necessarily imply a good evaluation scheme.\footnote{For example, in the "original" case, a very bad hypothetical evaluation scheme that aligns a random segment to the source can achieve the same 0\% NA ratio as groundtruth.} The correlation with human judgments should still be treated as the ultimate meta-evaluation metric.

\begin{table}[h]
    \centering
    \scalebox{0.75}{
        \begin{tabular}{@{}lrrr@{\hspace*{2pt}}l@{}}
            \toprule
                                          & \multicolumn{1}{c}{\textbf{COMET}} & \multicolumn{1}{c}{\textbf{MetricX}} & \multicolumn{2}{c}{\textbf{NA Ratio ($\left|\Delta_\text{Gold}\right|$)}} \\ \midrule
            \textbf{Original}             & \multicolumn{1}{l}{}               & \multicolumn{1}{l}{}                 & \multicolumn{1}{l}{}                 &                                     \\
            ersatz+LASER                  & 0.3072                             & 0.3051                               & \quad\quad\quad 1.2\%                & (1.2\%)                             \\
            ersatz+BGE-m3                 & 0.3037                             & 0.3055                               & 1.3\%                                & (1.3\%)                             \\
            ersatz+BGE-m3-ft              & \textbf{0.3085}                    & \textbf{0.3074}                      & \textbf{0.7\%}                       & \textbf{(0.7\%)}                    \\
            spacy+BGE-m3-ft               & 0.3028                             & 0.3049                               & 1.4\%                                & (1.4\%)                             \\ \midrule
            \textbf{Over-Translate}       & \multicolumn{1}{l}{}               & \multicolumn{1}{l}{}                 & \multicolumn{1}{l}{}                 &                                     \\
            ersatz+LASER                  & 0.3331                             & 0.3386                               & 8.6\%                                & (1.4\%)                             \\
            ersatz+BGE-m3                 & 0.3233                             & 0.3252                               & 9.8\%                                & (0.2\%)                             \\
            ersatz+BGE-m3-ft              & 0.3532                             & 0.3499                               & 11.2\%                               & (1.2\%)                             \\
            spacy+BGE-m3-ft               & \textbf{0.3554}                    & \textbf{0.3505}                      & \textbf{10.1\%}                      & \textbf{(0.1\%)}                    \\ \midrule
            \textbf{Under-Translate}      & \multicolumn{1}{l}{}               & \multicolumn{1}{l}{}                 & \multicolumn{1}{l}{}                 &                                     \\
            ersatz+LASER                  & 0.3405                             & 0.3176                               & \textbf{6.3\%}                       & \textbf{(3.7\%)}                    \\
            ersatz+BGE-m3                 & 0.3381                             & 0.3176                               & 4.2\%                                & (5.8\%)                             \\
            ersatz+BGE-m3-ft              & \textbf{0.3493}                    & \textbf{0.3268}                      & 5.8\%                                & (4.2\%)                             \\
            spacy+BGE-m3-ft               & 0.3481                             & 0.3258                               & 6.0\%                                & (4.0\%)                             \\ \midrule
            \textbf{Flex-Boundary}        & \multicolumn{1}{l}{}               & \multicolumn{1}{l}{}                 & \multicolumn{1}{l}{}                 &                                     \\
            ersatz+LASER                  & 0.3041                             & 0.3017                               & 1.9\%                                & (1.9\%)                             \\
            ersatz+BGE-m3                 & 0.3001                             & 0.3012                               & 2.2\%                                & (2.2\%)                             \\
            ersatz+BGE-m3-ft              & \textbf{0.3067}                    & \textbf{0.3033}                      & \textbf{1.2\%}                       & \textbf{(1.2\%)}                    \\
            spacy+BGE-m3-ft               & 0.3066                             & \textbf{0.3033}                      & 1.5\%                                & (1.5\%)                             \\ \bottomrule
        \end{tabular}
    }
    \caption{Ablation study on different sentence embeddings and segmenters.
        Numbers are calculated similarly to Table \ref{tab:main} but only include reference-based metrics due to space limits.
        Boldface numbers indicate the highest correlation for the first two columns, and the NA ratio with the smallest $\left|\Delta_\text{Gold}\right|$ for the last column.
    }
    \vspace{-0.279cm}
    \label{tab:abl}
\end{table}

\subsection{Results}
\paragraph{Main Results}
Table \ref{tab:main} shows a concise version of our main results (averaged across three language pairs).
In terms of correlation with human judgments, SEGALE achieves near-ideal performance, outperforming mwerSegmenter while maintaining comparable correlation with human judgments to Gold.
The gap is especially significant for the "under-translate" case, where mwerSegmenter suffers significant performance drops while SEGALE remains robust.
For a detailed version with per-language-pair breakdown, please refer to Appendix~\ref{app:supplementary_results}.
The readers shall see that the trend shown in Table \ref{tab:main} is generally consistent across all language pairs.

For NA ratio, we can observe that mwerSegmenter perfectly matches the groundtruth in two settings ("original" and "flex-boundary").
However, upon closer inspection, we conclude that this is not because mwerSegmenter can detect over-/under-translation errors more accurately, but rather because mwerSegmenter was not designed to account for certain translation anomalies.
For example, one crucial problem with mwerSegmenter is that it does not have an explicit mechanism to handle null alignments at the sentence level, nor does it have the semantic features that allows it to flag semantically irrelevant sentence pairs.
Hence, in the event of a system generating a hallucinated sentence, mwerSegmenter simply merges it to an arbitrary neighboring sentence, which can cause the underlying metric to miss over-translation errors (shown in Section \ref{sec:preliminaries}).
On the other hand, the mwerSegmenter version we experimented with does not allow for empty translation hypotheses, which can cause chunks of other well-translated sentences to be chopped up as an attempt to minimize deletion errors.
Obviously, this also greatly interferes with the scores generated by the underlying metric.
As for SEGALE, while it is also not perfect with inferring null alignments, the deviations from the groundtruth are more modest and are less likely to lead to catastrophic performance drops like mwerSegmenter in the "under-translate" case.

While mwerSegmenter doesn't provide sufficient logging that would allow us to pinpoint the exact reason for its performance drop, looking at the segmented text,
it is clear that mwerSegmenter's algorithm struggles to handle synonym substitution and paraphrasing, particularly when translation quality is poor and significant structural changes are present.
This exemplifies the limitation of using WER, instead of a semantic-based score, as the scoring function for segmentation and alignment.

The readers should also note that while the performance trend remains consistent between reference-based and reference-free metrics in our experiments, mwerSegmenter does require a reference translation as input, which in practice limits its applicability in reference-free evaluation settings.
SEGALE, on the other hand, directly performs cross-lingual sentence alignment without requiring a reference translation.

\paragraph{Impact of Sentence Embedding}
Table \ref{tab:abl} shows a comparison of SEGALE with different sentence embeddings.
It can be observed that our fine-tuned BGE-M3 embedding consistently outperforms LASER and the original BGE-M3 embedding in all data configurations.
The benefits of fine-tuning are especially significant for the "over-translate" and "under-translate" cases, which shows that our fine-tuning process successfully specializes the embedding model for capturing over- and under-translation errors during the coarse-to-fine search process of the sentence alignment step.

\paragraph{Impact of Sentence Segmenter}
Most of the numbers reported in this paper are calculated with the ersatz sentence segmenter.
We also experimented with spaCy as the sentence segmenter as another ablation study, also shown in Table \ref{tab:abl}.
Although we observed that spaCy tends to segment sentences into smaller units than ersatz (which does not align well with the long segments in WMT test sets), the impact on the performance seems to be minimal as we did not observe a consistent trend.

\section{Evaluation of Book-Length Translation Capability of Existing LLMs}

Now that we have validated our evaluation method on WMT 2024 metrics shared task dataset,
we briefly demonstrate that our method can be applied to assess the book-length translation capability of existing LLMs by conducting a similar experiment as~\citet{wang-etal-2024-benchmarking}.
Our dataset comes from the Chinese-English (zh-en) section of the WMT 2024 Discourse-Level Literary Translation task \cite{wang-etal-2024-findings}.
Because the test set only contains book chapters instead of full books, we randomly pick a book with ID \texttt{2-xzltq} from the training split of the dataset and use it as our test set.

\begin{table}[t]
    \centering
    \scalebox{0.72}{
    \begin{tabular}{ll}
    \toprule
    \textbf{Model ID} & \textbf{Reference} \\
    \midrule
    \texttt{utter-project/EuroLLM-9B-Instruct} & \citet{eurollm} \\
    \texttt{Qwen/Qwen2.5-14B-Instruct} & \citet{qwen2-5} \\
    \texttt{Qwen/Qwen2.5-72B-Instruct} & \citet{qwen2-5} \\
    \texttt{meta-llama/Llama-3.1-8B-Instruct} & \citet{llama-3} \\
    \texttt{meta-llama/Llama-3.1-70B-Instruct} & \citet{llama-3} \\
    \texttt{CohereForAI/aya-expanse-8b} & \citet{aya-expanse} \\
    \texttt{CohereForAI/aya-expanse-32b} & \citet{aya-expanse} \\
    \bottomrule
    \end{tabular}
    }
    \caption{List of LLMs evaluated for book-length translation capability.}
    \label{tab:model_list}
    \end{table}

The LLMs evaluated are listed in Table~\ref{tab:model_list}.
For simplicity, we adopted the same prompt and translation extraction procedure as used in WMT 2024 general machine translation shared task\footnote{\url{https://github.com/wmt-conference/wmt-collect-translations/blob/704b3825730f93a3ee3a0fda44af9414937b6d5a/tools/prompts.py\#L23}} for all the LLMs.
Since current LLMs are constrained by maximum generation lengths and cannot translate the entire book in a single pass, we divide the content into segments of 1k, 2k, 4k, and 8k tokens, using tokenization from the tiktoken tokenizer\footnote{\url{https://github.com/openai/tiktoken}}\hochkomma\footnote{Note that tiktoken tokenizer is only used to count the number of tokens in the segments and LLM-specific tokenization will still be applied during inference.}.
Most of these models have a maximum generation length of 8k tokens, except for EuroLLM, which is capped at 4k.

\begin{figure}[t]
    \centering
    \begin{subfigure}[htbp]{0.48\textwidth}
        \centering
        \includegraphics[width=\textwidth]{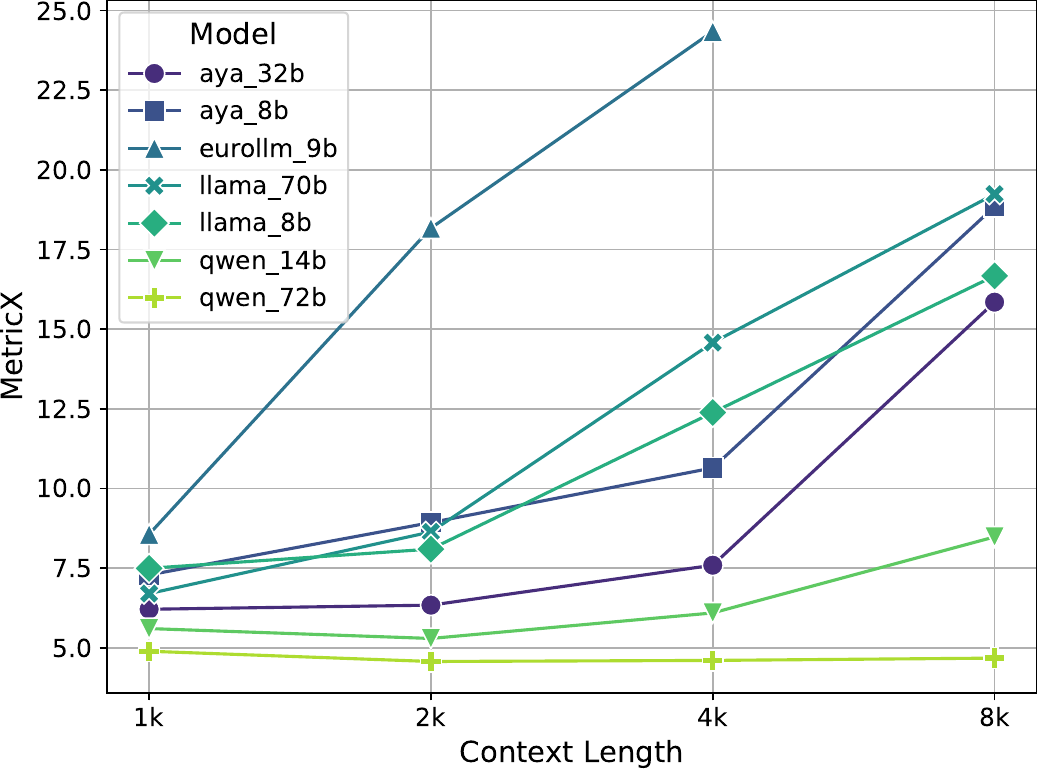}
        \caption{Translation Quality (SEGALE-MetricX $\downarrow$)}
        \label{fig:llm_cap_metricx}
    \end{subfigure}
    \begin{subfigure}[htbp]{0.48\textwidth}
        \centering
        \includegraphics[width=\textwidth]{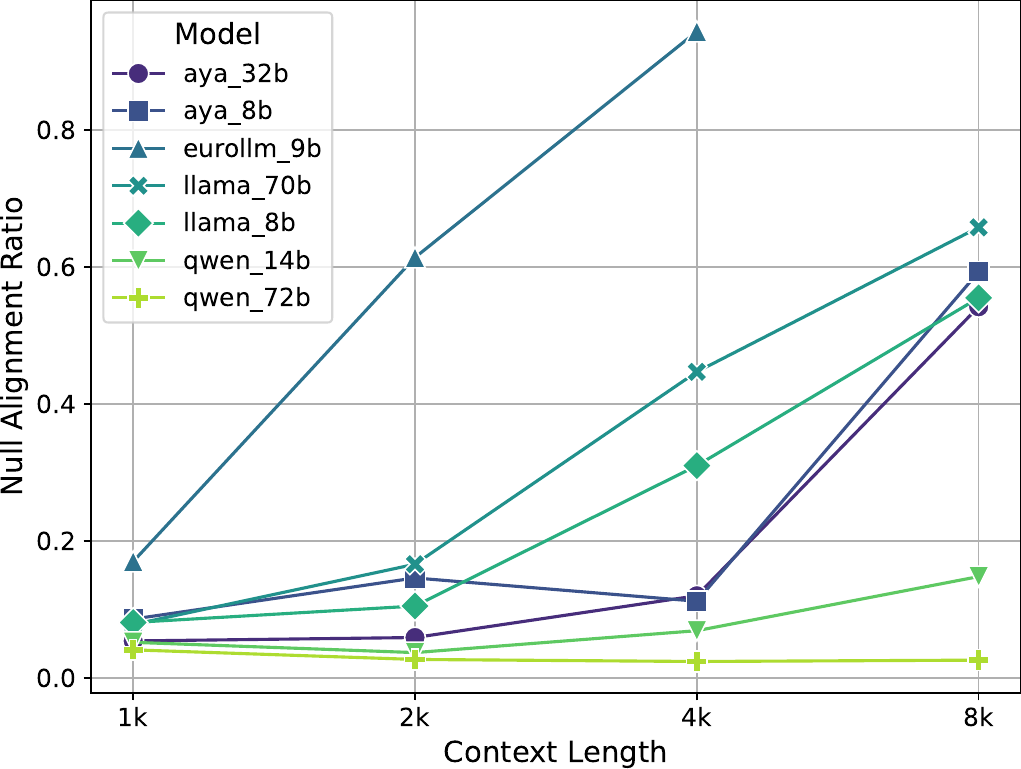}
        \caption{Null Alignment Ratio ($\downarrow$)}
        \label{fig:llm_cap_nar}
    \end{subfigure}
    \caption{LLM Translation Performance at Different Context Lengths}
    \label{fig:llm_cap}
\end{figure}

Figure \ref{fig:llm_cap} shows the translation quality and NA ratio of the LLMs at different context lengths.
Most models exhibit a sharp degradation in translation quality at context length of 4k or 8k.
For example, at 4k context length, EuroLLM refuses to translate as instructed, but rather resorts to summarizing the input document in the target language.
Comparing with the trend in NA ratio, it is also clear that under-translation/over-translation errors played a significant role in such degradation.
The only noteworthy exception is \texttt{Qwen2.5-72B-Instruct}, which shows a much more stable performance across different context lengths.
In fact, with the increasing context length up to 4k, there is a small improvement in translation quality, which shows its ability to utilize long-context information to obtain better translations.

This benchmark shows a significant gap between claimed max generation length and the actual capability of LLMs to translate long-context documents.
As future work keeps improving LLM's long-context processing capabilities, we call on the community to adopt this evaluation practice to gain better insights into such capabilities in downstream applications such as machine translation.

\section{Conclusion}

We propose SEGALE, a novel extension scheme that enables evaluation metrics to evaluate unsegmented document-level translations of arbitrary lengths.
SEGALE works with any existing evaluation metric and eliminates the dependency on sentence-level prompting and pre-segmented reference translations.
Experimental results show that our extension scheme achieves strong correlation with human judgments while demonstrating robustness to common LLM translation anomalies like over- and under-translation.
Through this work, we aim to facilitate machine translation research in its ongoing shift away from sentence-level paradigm, while also offering new perspectives for evaluating LLMs' long-context generation capabilities.

\section*{Limitations}

\subsection*{Underlying Metrics}

We acknowledge that an LLM-based metric based on long-context, open-source LLMs is a promising (and probably the eventual) solution to the problem of long-context MT evaluation.
While previous work has shown that LLM-based metrics such as GEMBA \cite{kocmi-federmann-2023-large} or AutoMQM \cite{fernandes-etal-2023-devil} can perform on-par as state-of-the-art BERT-based metrics such as COMET, they have to rely on GPT-4 or GPT-4o as the underlying LLM and are currently prohibitively expensive for MT evaluation of book-length documents.
We plan to explore the potential of the open-source LLM counterparts of these metrics for long-context MT evaluation in a separate study.


All underlying metrics explored in our experiments are sentence-level and do not explicitly incorporate discourse-level information.
There are existing methods that extend sentence-level metrics with surrounding context, such as \citet{vernikos-etal-2022-embarrassingly} and \citet{raunak-etal-2024-slide}.
We did not choose to go down that path because it will further complicate our experimental setup, and adding extra context to the underlying metrics is an orthogonal problem to our proposed extension scheme.
It is worth noting that the modular nature of our proposed extension scheme makes it fully compatible with these extensions, and we expect that such combination will yield better results.

The current design of our adaptive penalty search strategy is built upon the observation that existing metrics have limited sensitivity to over- and under-translation errors (Section \ref{sec:preliminaries}).
Right now, such errors are penalized by a fixed penalty over null alignments, which could appear crude as the underlying metrics continue to improve.
In such case, the hyperparameters of the adaptive penalty search strategy may need to be revisited to delegate more of these errors to the underlying metrics.
The downside of this approach is that the NA ratio may become a less reliable indicator of over- and under-translation errors.

\subsection*{Meta-Evaluation and Experimental Setup}

We did not include a direct comparison to other sentence alignment methods like Bleualign~\citep{sennrich-volk-2010-mt}.
We focus on Vecalign, whose superior performance to Bleualign is well-documented \cite{thompson-koehn-2019-vecalign,steingrimsson-etal-2023-sentalign}.
Besides, the adaptive penalty search mechanism is an enhancement designed specifically for Vecalign and is crucial for our approach's effectiveness.
This mechanism is incompatible with other aligners, which lack the fine-grained control over null alignments that is required by adaptive penalty search.

Another aspect worth considering is that the human judgments used in our evaluation do not explicitly instruct the annotators to consider discourse-level information.
Hence, whether those extra information will show benefits in meta-evaluations based on current human judgments remains unclear.
Since WMT 2025\footnote{\url{https://www2.statmt.org/wmt25/translation-task.html}} has recently shifted to include multi-line, long-form texts in its human evaluation setup, exploring discourse-level effects would be more meaningful using this updated dataset.

Our synthetic evaluation setup approximates real-world LLM translations but does not capture behaviors such as discourse-level content reordering.
We plan to conduct more extensive real-world testing of our method and compare it against a broader range of document-level MT evaluation paradigms in future work.


\subsection*{Fine-tuned Embedding Model}

The fine-tuned BGE-m3 embedding model used in our proposed extension scheme is largely a proof-of-concept, due to the fact that it is trained on a small dataset, covering only languages of our interest.
We believe that a specialized text embedding model for sentence alignment is not only useful for our proposed extension scheme, but also for its more traditional use cases, such as curation of web-crawled data.
In the future, we plan to explore extending the volume of the training data and supported languages to improve the usability of our proposed extension scheme.



\section*{Acknowledgments}

We thank Brian Thompson for discussions regarding Vecalign, Rachel Wicks for discussions regarding ersatz, Miguel Romero Calvo for discussions regarding contrastive fine-tuning of embedding models, and Oleksii Hrinchuk for discussions regarding mwerSegmenter.
We also thank Simeng Sun, Elizabeth Salesky, and anonymous reviewers for their valuable comments and suggestions on earlier drafts of this paper.
This project has been supported by NVIDIA Taiwan Research \& Dev Center (TRDC) Grant.

\bibliography{anthology_split_250224_no_editor,custom}

\appendix
\clearpage
\section{Synthetic Test Data Creation}\label{app:merge_case}

To evaluate the robustness of our evaluation framework, we construct synthetic test data simulating three common alignment challenges: under-/over-translation and varied sentence boundaries.
\subsection{Synthetic Under- and Over-Translations}
We simulate under- and over-translation scenarios by randomly removing 10\% of the segments from either the source/reference sides or the system translations, respectively. To maintain meaningful context, we avoid sampling from documents that contain only a single segment.

\subsection{Synthetic Sentence Boundary Variation}
To simulate sentence boundary variation, we generate synthetic test data by merging 10\% of adjacent segments in the source side. This process is conducted using GPT-4o, with the following constraints:
\begin{itemize}
    \item Only segments that are neither the first nor the last in a document are eligible for merging.
    \item For each eligible candidate, an attempt is made to merge it with its subsequent segment.
    \item Merging is only accepted if the semantic difference between the merged and original segments is minimal. We use BLEURT~\citep{sellam2020bleurt} to assess the semantic similarity, accepting only merges with a BLEURT score greater than 0.85. If a candidate pair fails this criterion, another eligible pair is sampled.
\end{itemize}

GPT-4o is instructed to merge adjacent segments into a single, fluent sentence without changing the original meaning, vocabulary, or the order of information. Figure~\ref{fig:merge_prompt} shows the prompt template used to guide the model. 

Figure \ref{fig:merge_result} provides an example of the merging process before and after GPT-4o rewriting. The sentences initially presented separately are transformed into a single sentence using appropriate transitional phrases.

This procedure enables us to test our evaluation method’s robustness in conditions reflecting realistic variations in sentence boundary alignments while ensuring that human annotations remain valid and can be directly reused without recalibration.

\begin{figure*}[h!]
\centering
\includegraphics[width=\textwidth]{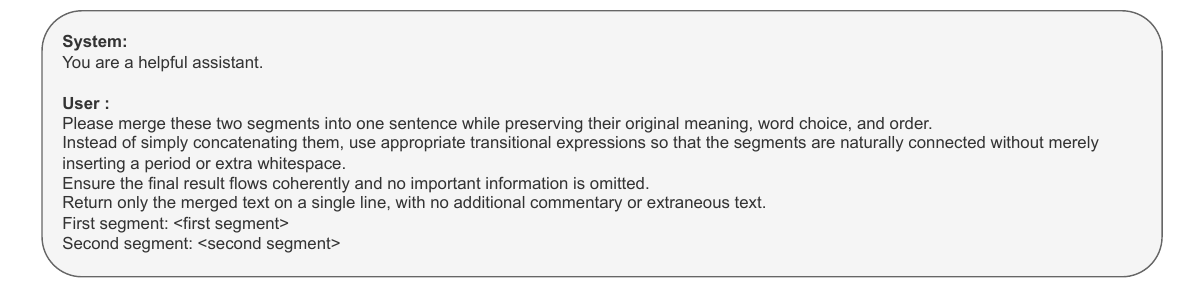}
\caption{Prompt used for instructing GPT-4o to merge sentences.}
\label{fig:merge_prompt}
\end{figure*}

\begin{figure*}[h!]
\centering
\includegraphics[width=\textwidth]{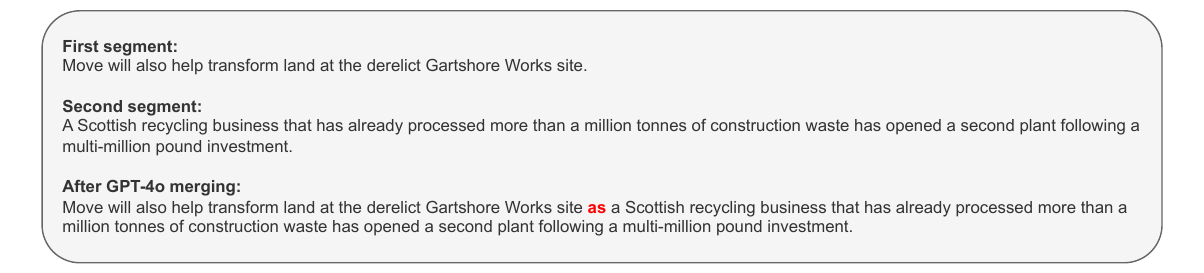}
\caption{Example of merged sentences before and after GPT-4o rewriting.}
\label{fig:merge_result}
\end{figure*}

\section{Implementation Details}\label{app:imple_details}
In this section, we provide detailed implementation information on extending existing evaluation metrics to book-length documents. 
Our proposed approach is designed as a flexible framework where the underlying models can be readily substituted. 
Here, we specifically outline the experimental configurations used in this study.


\subsection{Sentence Segmentation}
For sentence segmentation, our experiments employ two different models:  \texttt{spaCy} and \texttt{ersatz}. 
\texttt{spaCy} requires specification of the target language, whereas \texttt{ersatz} is language-agnostic, making it suitable for multilingual segmentation tasks.

The experiments in this paper cover five languages: English, German, Spanish, Japanese, and Chinese. Corresponding \texttt{spaCy} models for each language are as follows:
\begin{itemize} \itemsep -2pt
    \item English (en): \texttt{en\_core\_web\_sm}
    \item German (de): \texttt{de\_core\_news\_sm}
    \item Spanish (es): \texttt{es\_core\_news\_sm}
    \item Japanese (ja): \texttt{ja\_core\_news\_sm}
    \item Chinese (zh): \texttt{zh\_core\_web\_sm}
\end{itemize}


\subsection{Sentence Alignment}\label{app:subsec:align_imp}
We adopt a robust sentence alignment strategy based on Vecalign~\citep{thompson-koehn-2019-vecalign}, leveraging multilingual sentence embeddings and an efficient dynamic programming approximation to identify many-to-many alignments between sentence segments.

In our experiments, we set the maximum number of allowed overlap size to 16. This allows us to search for source-target sentence alignments of size \( N\text{--}M \), where \( N + M \leq 16 \). 
While we use a fixed overlap size in our experiments, it can be estimated from reference documents. Detailed explanations are provided in Appendix~\ref{app:est_overlap}.

\paragraph{Adaptive Penalty Search.}  
In Vecalign, null alignments are handled via a skip cost, parameterized by a percentile-based threshold \( \beta_{\text{skip}} \), representing a quantile in the empirical distribution of 1-1 alignment costs. 
To identify the optimal \( \beta_{\text{skip}} \), we perform a search starting from 0.2, which is the default value in Vecalign, and progressively decrease it in small steps of 0.005. At each step, we detect signals of over-deletion. Upon detection, we revert to the previous step and treat it as the final alignment result.
\paragraph{Heuristic Termination Conditions.}  
The search is terminated based on two heuristic signals that indicate over-deletion:
\begin{itemize}
    \item[(a)] The average alignment cost falls below 0.3.
    \item[(b)] The null alignment ratio at a step exceeds 0.15.
\end{itemize}
Both patterns suggest that the skip cost has become too lenient, leading to excessive null alignments.
In addition, two rare edge cases are handled with early stopping:
\begin{itemize}
    \item[(a)] If the average alignment cost increases rather than decreases at a step, indicating misalignments with semantically distant content.
    \item[(b)] If the average alignment cost exceeds 0.7, indicating poor alignment quality overall.
\end{itemize}

\paragraph{Parameter Tuning.}  
These heuristic termination conditions were tuned empirically on the test set from the WMT24 Discourse-Level Literary Translation shared task, using only the Chinese$\rightarrow$English portion. 
To remain within the context length of our selected LLMs, we segment each instance from the training and validation sets into chunks of up to 1024 tokens. Translations are generated using \texttt{meta-llama/Llama-3.1-8B-Instruct} and \texttt{GPT-4o}$_{\texttt{~2024-08-06}}$, and sentence embeddings are computed using LASER. 
We empirically validate alignment quality by manually inspecting whether translation errors are consistently marked as null alignments. Once verified, the same configuration is used for all experiments in this paper.

While the heuristic termination conditions are robust in our experiments, they may vary depending on the translation direction and source/reference sentence boundaries. 
These parameters can be further refined using reference translations, which are typically assumed to be perfectly aligned -- i.e., with a null alignment (NA) ratio of zero. 
This provides a basis for estimating how much the NA ratio increases when the skip cost becomes too lenient, as well as the expected alignment cost under ideal conditions. 
These estimates can then inform the selection of appropriate heuristic parameters when evaluating future system translations.

\subsection{Details for Text Embedding Model Fine-Tuning}\label{subsec:emb_finetune}

We used News Commentary 18.1 parallel data from any language pairs that is a combination of Chinese (zh), English (en), German (de), Japanese (ja), and Spanish (es), which are the languages of interest in our evaluation.
For each language pair, we use either the full dataset or only first 10,000 lines of the dataset, whichever is smaller.
We build the example triplets with the following: we concatenate each of the two neighboring sentence pairs in the parallel corpus and use the source/target side as the query/positive example, respectively.
As for the negative example, we always construct two variants: (1) we randomly drop one of the two sentences on the target side (2) we retrieve a nearby sentence by randomly looking forward 1 to 3 sentences in the dataset and use it to substitute the second sentence on the target side.
The intuition behind these example triplets is for the model to better distinguish a good translation from (1) an incomplete translation, and (2) a sentence that has a similar topic but is not a translation.
The resulting synthetic dataset has 130,436 examples.

We fine-tune the BGE-M3 embedding on the synthetic dataset with InfoNCE loss \cite{infonce} for two epochs.
We did not conduct an extensive hyperparameter search, but simply use the setup in the fine-tuning tutorial in the FlagEmbedding package\footnote{\url{https://github.com/FlagOpen/FlagEmbedding/blob/024e789d599eb4cf9a208e98d27508ad455f5ecb/Tutorials/7_Fine-tuning/7.1.2_Fine-tune.ipynb}}.
We directly used the checkpoint at the end of the training without using a validation set to select the best checkpoint.

\subsection{Setting Overlap Size for Alignment}\label{app:est_overlap}
Beyond the text embedding model, we also observe that the choice of overlap size in Vecalign can significantly impact the alignment quality.
The overlap size defines the size of the blocks that are compared to each other in the alignment search.
A small overlap size causes some ideal alignments to fall out of search space.
For example, if the overlap size is set to 8, but a long source sentence should be aligned to a sentence block of size 16 on the target side, such groundtruth alignment will never be considered by the search algorithm.
On the other hand, a large overlap size will increase the computational cost.

With datasets that come with human-segmented sentence boundaries and alignment (which covers the vast majority of use cases), one can easily and accurately estimate the appropriate overlap size by re-segmenting the reference documents with sentence segmenter, calculate the maximum ratio of sentences as segmented by human and by the sentence segmenter.
With datasets that does not come with human-segmented sentence boundaries or references, one would have to first conduct a pilot study with sentence-level translation to estimate the appropriate overlap size.
However, the good news is that with a given sentence segmenter, the overlap size only needs to be estimated once per language, and can be reused for different datasets.
Besides, in the case where estimation is not very accurate, one can always err on the safe side and set a larger overlap size.

\subsection{Details for Evaluation Setup}
For budgetary reasons, WMT datasets from recent years exclude some segments from MQM annotations.
To ensure meaningful comparison between the automatic metrics and human judgments at the document level, we exclude documents from the evaluation if more than 20\% of the segments are missing MQM annotations.

Running mwerSegmenter requires a tokenization preprocessing step.
For ja-zh, we follow the anonymous reviewer's recommendation to tokenize the output into individual characters.
For other languages, we use the tokenizer implemented in Sacremoses\footnote{\url{https://pypi.org/project/sacremoses/}}.

\section{Supplementary Results} \label{app:supplementary_results}
Since the results in the main paper are condensed versions with averaging across different language pairs or showing only a subset of the metrics evaluated, we attach the full breakdown of the results in Table \ref{tab:full} for readers' reference.

\begin{sidewaystable*}[h]
\centering
    \scalebox{0.8}{
        \begin{tabular}{@{}lrrrrrrrrrrrrrrr@{}}
            \toprule
            \multicolumn{1}{c}{}   & \multicolumn{3}{c}{\textbf{COMET}}                                 & \multicolumn{3}{c}{\textbf{COMET-QE}}                                & \multicolumn{3}{c}{\textbf{METRICX}}                                 & \multicolumn{3}{c}{\textbf{METRICX-QE}}                              & \multicolumn{3}{c}{\textbf{NA Ratio}}                                  \\
            \multicolumn{1}{c}{}   & \multicolumn{1}{c}{\textbf{en-de}} & \multicolumn{1}{c}{\textbf{en-es}} & \multicolumn{1}{c}{\textbf{ja-zh}} & \multicolumn{1}{c}{\textbf{en-de}} & \multicolumn{1}{c}{\textbf{en-es}} & \multicolumn{1}{c}{\textbf{ja-zh}} & \multicolumn{1}{c}{\textbf{en-de}} & \multicolumn{1}{c}{\textbf{en-es}} & \multicolumn{1}{c}{\textbf{ja-zh}} & \multicolumn{1}{c}{\textbf{en-de}} & \multicolumn{1}{c}{\textbf{en-es}} & \multicolumn{1}{c}{\textbf{ja-zh}} & \multicolumn{1}{c}{\textbf{en-de}} & \multicolumn{1}{c}{\textbf{en-es}} & \multicolumn{1}{c}{\textbf{ja-zh}} \\ \midrule
            \textbf{Original}      & \multicolumn{1}{l}{}               & \multicolumn{1}{l}{}               & \multicolumn{1}{l}{}               & \multicolumn{1}{l}{}               & \multicolumn{1}{l}{}               & \multicolumn{1}{l}{}               & \multicolumn{1}{l}{}               & \multicolumn{1}{l}{}               & \multicolumn{1}{l}{}               & \multicolumn{1}{l}{}               & \multicolumn{1}{l}{}               & \multicolumn{1}{l}{}               & \multicolumn{1}{l}{}               & \multicolumn{1}{l}{}               & \multicolumn{1}{l}{}               \\
            Gold                   & 0.3239                             & 0.2320                             & 0.3770                             & 0.2667                             & 0.2248                             & 0.3486                             & 0.3468                             & 0.2352                             & 0.3434                             & 0.2807                             & 0.1995                             & 0.3247                             & 0\%                                & 0\%                                & 0\%                                \\
            mwerSegmenter          & 0.2551                             & 0.2376                             & 0.3696                             & 0.2045                             & 0.2228                             & 0.3368                             & 0.2666                             & 0.2413                             & 0.3318                             & 0.2009                             & 0.2050                             & 0.3022                             & 0\%                                & 0\%                                & 0\%                                \\
            SEGALE-ersatz-BGE-M3-ft  & 0.3229                             & 0.2410                             & 0.3616                             & 0.2662                             & 0.2291                             & 0.3351                             & 0.3458                             & 0.2454                             & 0.3310                             & 0.2776                             & 0.2050                             & 0.3063                             & 0.25\%                             & 0.75\%                             & 1.32\%                             \\
            SEGALE-spaCy-BGE-M3-ft   & 0.3222                             & 0.2350                             & 0.3513                             & 0.2652                             & 0.2261                             & 0.3314                             & 0.3457                             & 0.2401                             & 0.3288                             & 0.2775                             & 0.2035                             & 0.3047                             & 0.99\%                             & 0.81\%                             & 2.34\%                             \\
            SEGALE-ersatz-LASER    & 0.3208                             & 0.2340                             & 0.3669                             & 0.2637                             & 0.2258                             & 0.3380                             & 0.3452                             & 0.2392                             & 0.3310                             & 0.2782                             & 0.2029                             & 0.3069                             & 0.23\%                             & 0.82\%                             & 2.40\%                             \\
            SEGALE-ersatz-BGE-M3   & 0.3193                             & 0.2300                             & 0.3617                             & 0.2635                             & 0.2205                             & 0.3344                             & 0.3448                             & 0.2339                             & 0.3377                             & 0.2779                             & 0.1977                             & 0.3108                             & 0.36\%                             & 0.81\%                             & 2.84\%                             \\
                                   & \multicolumn{1}{l}{}               & \multicolumn{1}{l}{}               & \multicolumn{1}{l}{}               & \multicolumn{1}{l}{}               & \multicolumn{1}{l}{}               & \multicolumn{1}{l}{}               & \multicolumn{1}{l}{}               & \multicolumn{1}{l}{}               & \multicolumn{1}{l}{}               & \multicolumn{1}{l}{}               & \multicolumn{1}{l}{}               & \multicolumn{1}{l}{}               & \multicolumn{1}{l}{}               & \multicolumn{1}{l}{}               & \multicolumn{1}{l}{}               \\
            \textbf{Over-Translate}  & \multicolumn{1}{l}{}               & \multicolumn{1}{l}{}               & \multicolumn{1}{l}{}               & \multicolumn{1}{l}{}               & \multicolumn{1}{l}{}               & \multicolumn{1}{l}{}               & \multicolumn{1}{l}{}               & \multicolumn{1}{l}{}               & \multicolumn{1}{l}{}               & \multicolumn{1}{l}{}               & \multicolumn{1}{l}{}               & \multicolumn{1}{l}{}               & \multicolumn{1}{l}{}               & \multicolumn{1}{l}{}               & \multicolumn{1}{l}{}               \\
            Gold                   & 0.3590                             & 0.3975                             & 0.3979                             & 0.3145                             & 0.3962                             & 0.3533                             & 0.3727                             & 0.3430                             & 0.3637                             & 0.3004                             & 0.2620                             & 0.3380                             & 10\%                               & 10\%                               & 10\%                               \\
            mwerSegmenter          & 0.2799                             & 0.3816                             & 0.3904                             & 0.2410                             & 0.3747                             & 0.3483                             & 0.2704                             & 0.3533                             & 0.3515                             & 0.2020                             & 0.2957                             & 0.3211                             & 10\%                               & 10\%                               & 10\%                               \\
            SEGALE-ersatz-BGE-M3-ft  & 0.3318                             & 0.3488                             & 0.3789                             & 0.3012                             & 0.3628                             & 0.3464                             & 0.3588                             & 0.3400                             & 0.3508                             & 0.2973                             & 0.3003                             & 0.3269                             & 9.29\%                             & 13.28\%                            & 11.09\%                            \\
            SEGALE-spaCy-BGE-M3-ft   & 0.3383                             & 0.3495                             & 0.3783                             & 0.3007                             & 0.3625                             & 0.3446                             & 0.3618                             & 0.3405                             & 0.3492                             & 0.3002                             & 0.3005                             & 0.3255                             & 9.47\%                             & 9.55\%                             & 11.35\%                            \\
            SEGALE-ersatz-LASER    & 0.3327                             & 0.3146                             & 0.3521                             & 0.2994                             & 0.3361                             & 0.3129                             & 0.3584                             & 0.3212                             & 0.3361                             & 0.2956                             & 0.2752                             & 0.3126                             & 8.16\%                             & 9.07\%                             & 8.66\%                             \\
            SEGALE-ersatz-BGE-M3   & 0.3104                             & 0.3102                             & 0.3494                             & 0.2783                             & 0.3278                             & 0.3135                             & 0.3381                             & 0.3030                             & 0.3344                             & 0.2794                             & 0.2574                             & 0.3092                             & 9.26\%                             & 8.61\%                             & 11.37\%                            \\
                                   & \multicolumn{1}{l}{}               & \multicolumn{1}{l}{}               & \multicolumn{1}{l}{}               & \multicolumn{1}{l}{}               & \multicolumn{1}{l}{}               & \multicolumn{1}{l}{}               & \multicolumn{1}{l}{}               & \multicolumn{1}{l}{}               & \multicolumn{1}{l}{}               & \multicolumn{1}{l}{}               & \multicolumn{1}{l}{}               & \multicolumn{1}{l}{}               & \multicolumn{1}{l}{}               & \multicolumn{1}{l}{}               & \multicolumn{1}{l}{}               \\
            \textbf{Under-Translate} & \multicolumn{1}{l}{}               & \multicolumn{1}{l}{}               & \multicolumn{1}{l}{}               & \multicolumn{1}{l}{}               & \multicolumn{1}{l}{}               & \multicolumn{1}{l}{}               & \multicolumn{1}{l}{}               & \multicolumn{1}{l}{\textbf{}}      & \multicolumn{1}{l}{}               & \multicolumn{1}{l}{}               & \multicolumn{1}{l}{\textbf{}}      & \multicolumn{1}{l}{}               & \multicolumn{1}{l}{}               & \multicolumn{1}{l}{}               & \multicolumn{1}{l}{}               \\
            Gold                   & 0.3488                             & 0.3349                             & 0.3725                             & 0.3050                             & 0.3416                             & 0.3056                             & 0.3542                             & 0.2564                             & 0.3199                             & 0.2881                             & 0.2406                             & 0.3208                             & 10\%                               & 10\%                               & 10\%                               \\
            mwerSegmenter          & 0.1198                             & 0.1935                             & 0.3415                             & 0.0947                             & 0.1826                             & 0.2937                             & 0.1169                             & 0.1266                             & 0.2876                             & 0.0754                             & 0.0950                             & 0.2619                             & 2.65\%                             & 3.64\%                             & 0.02\%                             \\
            SEGALE-ersatz-BGE-M3-ft  & 0.3378                             & 0.3499                             & 0.3601                             & 0.3000                             & 0.3619                             & 0.3266                             & 0.3523                             & 0.2944                             & 0.3338                             & 0.2909                             & 0.2566                             & 0.3082                             & 5.10\%                             & 6.62\%                             & 5.72\%                             \\
            SEGALE-spaCy-BGE-M3-ft   & 0.3378                             & 0.3481                             & 0.3583                             & 0.2998                             & 0.3610                             & 0.3266                             & 0.3528                             & 0.2923                             & 0.3322                             & 0.2916                             & 0.2556                             & 0.3088                             & 5.47\%                             & 6.61\%                             & 6.01\%                             \\
            SEGALE-ersatz-LASER    & 0.3392                             & 0.3415                             & 0.3409                             & 0.2934                             & 0.3524                             & 0.3132                             & 0.3488                             & 0.2850                             & 0.3189                             & 0.2891                             & 0.2529                             & 0.2926                             & 5.55\%                             & 6.34\%                             & 7.13\%                             \\
            SEGALE-ersatz-BGE-M3   & 0.3412                             & 0.3225                             & 0.3507                             & 0.3001                             & 0.3441                             & 0.3142                             & 0.3504                             & 0.2726                             & 0.3299                             & 0.2903                             & 0.2447                             & 0.3042                             & 2.98\%                             & 4.14\%                             & 5.54\%                             \\
                                   & \multicolumn{1}{l}{}               & \multicolumn{1}{l}{}               & \multicolumn{1}{l}{}               & \multicolumn{1}{l}{}               & \multicolumn{1}{l}{}               & \multicolumn{1}{l}{}               & \multicolumn{1}{l}{}               & \multicolumn{1}{l}{}               & \multicolumn{1}{l}{}               & \multicolumn{1}{l}{}               & \multicolumn{1}{l}{}               & \multicolumn{1}{l}{}               & \multicolumn{1}{l}{}               & \multicolumn{1}{l}{}               & \multicolumn{1}{l}{}               \\
            \textbf{Flex-Boundary}   & \multicolumn{1}{l}{}               & \multicolumn{1}{l}{}               & \multicolumn{1}{l}{}               & \multicolumn{1}{l}{}               & \multicolumn{1}{l}{}               & \multicolumn{1}{l}{}               & \multicolumn{1}{l}{}               & \multicolumn{1}{l}{}               & \multicolumn{1}{l}{}               & \multicolumn{1}{l}{}               & \multicolumn{1}{l}{}               & \multicolumn{1}{l}{}               & \multicolumn{1}{l}{}               & \multicolumn{1}{l}{}               & \multicolumn{1}{l}{}               \\
            Gold                   & 0.3220                             & 0.2321                             & 0.3746                             & 0.2632                             & 0.2270                             & 0.3463                             & 0.3433                             & 0.2372                             & 0.3394                             & 0.2768                             & 0.2011                             & 0.3216                             & 0\%                                & 0\%                                & 0\%                                \\
            mwerSegmenter          & 0.2301                             & 0.1853                             & 0.3680                             & 0.1803                             & 0.1831                             & 0.3342                             & 0.2365                             & 0.1929                             & 0.3279                             & 0.1741                             & 0.1652                             & 0.2999                             & 0\%                                & 0\%                                & 0\%                                \\
            SEGALE-ersatz-BGE-M3-ft  & 0.3213                             & 0.2299                             & 0.3688                             & 0.2620                             & 0.2227                             & 0.3392                             & 0.3414                             & 0.2379                             & 0.3306                             & 0.2721                             & 0.2010                             & 0.3060                             & 0.98\%                             & 1.59\%                             & 1.05\%                             \\
            SEGALE-spaCy-BGE-M3-ft   & 0.3211                             & 0.2313                             & 0.3673                             & 0.2616                             & 0.2237                             & 0.3389                             & 0.3415                             & 0.2391                             & 0.3292                             & 0.2720                             & 0.2022                             & 0.3053                             & 1.43\%                             & 1.64\%                             & 1.35\%                             \\
            SEGALE-ersatz-LASER    & 0.3182                             & 0.2300                             & 0.3641                             & 0.2609                             & 0.2232                             & 0.3357                             & 0.3416                             & 0.2373                             & 0.3263                             & 0.2746                             & 0.2003                             & 0.3033                             & 1.31\%                             & 1.81\%                             & 2.55\%                             \\
            SEGALE-ersatz-BGE-M3   & 0.3169                             & 0.2280                             & 0.3554                             & 0.2596                             & 0.2183                             & 0.3312                             & 0.3406                             & 0.2339                             & 0.3292                             & 0.2729                             & 0.1971                             & 0.3035                             & 1.19\%                             & 2.03\%                             & 3.33\%                             \\ \bottomrule
        \end{tabular}  
    }
    \caption{Full breakdown of our results on the WMT 2024 Metrics Shared Task dataset.}
    \label{tab:full}
\end{sidewaystable*}

\section{Licensing of Artifacts}

Almost all code, model, and data artifacts we used in this paper are publicly available with permissive licenses (MIT/Apache 2.0/CC-BY-4.0).
The only exceptions are Aya models (CC-BY-NC 4.0), and GPT-4o (OpenAI API Terms of Use), which still allows research use.
We also plan to release all created code, model, and data artifacts under a permissive license.

\section{Use of AI Assistants}

We used a code editor with generative AI functionalities during code development and paper writing (in the latter case, it only assists with LaTeX code completion and minor text editing).
We also used various AI assistants for creating miscellaneous single-use data processing scripts, as well as all the figures in this paper.
All AI-generated artifacts were carefully reviewed and accepted by the authors.

\end{document}